\documentclass[10pt, a4paper]{article}
\usepackage{lrec2022} 
\usepackage[utf8]{inputenc}
\usepackage{multibib}
\newcites{languageresource}{Language Resources}
\usepackage{graphicx}
\usepackage{tabularx}

\usepackage{arabtex}
\usepackage{latexsym}

\usepackage{soul}
\usepackage{titlesec}
\titleformat{\section}{\normalfont\large\bfseries\center}{\thesection.}{1em}{}
\titleformat{\subsection}{\normalfont\SmallTitleFont\bfseries\raggedright}{\thesubsection.}{1em}{}
\titleformat{\subsubsection}{\normalfont\normalsize\bfseries\raggedright}{\thesubsubsection.}{1em}{}
\renewcommand\thesection{\arabic{section}}
\renewcommand\thesubsection{\thesection.\arabic{subsection}}
\renewcommand\thesubsubsection{\thesubsection.\arabic{subsubsection}}

\usepackage{epstopdf}
\usepackage[utf8]{inputenc}
\usepackage{arabtex}
\usepackage{utf8}
\usepackage{hyperref}
\usepackage{xstring}
\usepackage[dvipsnames]{xcolor}
\usepackage{color}
\usepackage{todonotes}

\usepackage{paralist}

\title{Towards Arabic Sentence Simplification via Classification and Generative Approaches  }

\name{Nouran Khallaf$^\dagger$$^\ddagger$, Serge Sharoff$^\dagger$} 

\address{ $^\dagger$University of Leeds, UK\\
$^\ddagger$Alexandria University, Egypt\\
$^\dagger${mlnak, s.sharoff}@leeds.ac.uk}

\abstract{
This paper presents an attempt to build a Modern Standard Arabic (MSA) sentence-level simplification system.  
We experimented with sentence simplification using two approaches: (i) a classification approach leading to lexical simplification pipelines which use Arabic-BERT, a pre-trained contextualised model, as well as a model of fastText word embeddings; and (ii) a generative approach, a Seq2Seq technique by applying a multilingual Text-to-Text Transfer Transformer mT5. We developed our training corpus by aligning the original and simplified sentences from the internationally acclaimed Arabic  novel  “Saaq  al-Bambuu”. We evaluate effectiveness of these methods by comparing the generated simple sentences to the target simple sentences using the BERTScore evaluation metric. The simple sentences produced by the mT5 model achieve P 0.72, R 0.68 and F-1 0.70 via BERTScore, while, combining Arabic-BERT and fastText achieves P 0.97, R 0.97 and F-1 0.97. In addition, we report a manual error analysis for these experiments.
 \\ 
 \newline 
 }

\begin{document}

\maketitleabstract

\section{Introduction}

Text Simplification (TS) is a Natural Language Processing (NLP) task that aims to reduce the linguistic complexity of the text while maintaining its meaning and original information \cite{saggion2017automatic,siddharthan2002architecture,collados2013syntactic}. \newcite{shardlow2014survey} stated that TS task may includes lexical or/and syntactic simplification to produce a new equivalent text which conveys the same meaning and message with simpler words and structure. According to this definition, TS involves text transformation with new lexical items and/or rewriting sentences to ensure both its readability and understandability for the target audience \cite{bott2011unsupervised}. This definition also suggests that TS could be classified as a type of Text Style Transfer (TST), where the target style of the generated text is “simple” \cite{jin2020deep}. Evidence suggests the importance of TS involves : (i) its usage in designing and simplifying the language curriculum for both second language and first language learners, in making text easy-to-read for first language early learners; in assisting first-language users with cognitive impairments and low literacy language level; (ii) being a fundamental pre-process in NLP applications such as text retrieval, extraction, summarization, categorization and translation \cite{saggion2017automatic}; and (iii) acting as a post-process step in Automatic speech recognition. 

Accordingly, the TS task varies depending on the final application or the target audience. Hence, there are various types of simplification systems based on the purpose and who is the end-user of the system. A reasonable approach to tackle this issue could be to follow a general simplification strategy. There are three key aspects of simple text that: (i) it is made up of frequent simple words, grammatically simple sentences, and direct language; (ii) unnecessary information is omitted ; (iii) it can be shorter by the number of words, but also with shorter sentences, which might lead to their increased number \cite{bott2011unsupervised,collados2013syntactic}. \newcite{collados2013syntactic} approached TS differently as he  came up with different opinion, that is a slightly simplified text for one user is generally simpler for any other users. But a more extensive simplification for a specific user, may lead to a more complex text for another user.

TS is an active NLP research area, like other ongoing research, its techniques have moved away from manually hand-crafted rules towards deep learning techniques \cite{sikka2020survey,al2021automated}. Most of these techniques were borrowed from closely related NLP tasks such as Machine Translation \cite{sikka2020survey} . This has influenced our experiments to demonstrate the effectiveness of two different methods to address the sentence simplification (SS) task as follows:
\paragraph{\emph{(1) Classification Approach}} SS is considered as a classification task that requires a decision on which word to replace or syntactic structure to regenerate in each complex sentence. This approach allows the application of the Lexical Simplification (LS) task pipeline, i.e that aims to control the readability attribute of the text and make it more accessible to different readers with various intellectual abilities. LS particularly involves word change, thus we experiment the effect of different  embedding representation on word classification decision. This approach highlights the impact of how the text is simplified either by applying word embedding, or contextualised embedding such as BERT \cite{devlin2018bert}.

\paragraph{\emph{(2) Generative Approach}} SS is considered as a translation task, in which the translation is done within the same language from a complex sentence as the source to a simplified sentence as the target \cite{zhu2010monolingual}. According to this perspective, SS generative model could be implemented using Machine Translation (MT) and monolingual text-to-text generation techniques. Thus, we combined all SS steps into one process which learns from the complex sentence how to generate the simple version. For this purpose, we applied a BERT-like pre-trained transformer to perform a sequence-to-sequence (Seq2Seq) algorithm.

The main contribution of this paper is to examine  different approaches for Arabic sentence simplification task using automatic and manual evaluation. To our knowledge, this is the first available Arabic sentence-level simplification system.\footnote{\url{https://github.com/Nouran-Khallaf/Lexical_Simplification}}

\section{Corpus and Tools}
The corpus used for training is a set of complex/simple parallel sentences that have been compiled from the internationally acclaimed Arabic novel “Saaq al-Bambuu” which has an authorized simplified version for students of Arabic as a second language \cite{al2016saud}. We assume that a successful sentence simplifier should be able to detect word/sentences in the original text that require simplification and simplify them in such a way as the original simple counterpart. The dataset consists of 2980 parallel sentences as illustrated in Table \ref{tabCorpus} and classified according to The Common European Framework of language proficiency Reference (CEFR)\footnote{\url{https://www.cambridgeenglish.org/exams-and-tests/cefr/}} .i.e is an international standard for describing language ability ranging from A1, A2 … up to   C2.
\begin{table}[!h]
\begin{center}
\begin{tabularx}{\columnwidth}{|l|X|X|}

      \hline
      \textbf{Levels}& \textbf{Sentence} & \textbf{Tokens}\\
      \hline
      Simple A+B &2980& 34447\\
      \hline
      Complex C & 2980 & 46521\\
      \hline
     \textbf{Total} & \textbf{5690} & 80968\\
      \hline

\end{tabularx}
\caption{Number of Sentences and Tokens available per each CEFR Level in Saaq al-Bambuu parallel corpus}
\label{tabCorpus}
 \end{center}
\end{table}

We aligned the words in the parallel “Saaq al-Bambuu” sentences using Eflomal\footnote{\small{\url{https://github.com/robertostling/eflomal}}}  word aligning tool that uses a Bayesian
model with Markov Chain Monte Carlo (MCMC) inference \cite{Ostling2016efmaral}. After aligning the words, we automatically identified four basic simplification types on word-level and sentence-level \cite{alva2017learning}, then annotate these types with the following labels :
\begin{itemize}
\item Deletions, DELETE (D) in the complex sentence. [word-level]
\item	Additions, ADD (A) in the simplified sentence. [sentence-level]
\item	Substitutions, REPLACE (R), a word in the complex sentence is replaced by a new word in the simplified sentence. [word-level]
\item	Rewrites, REWRITE (RW) words shared in both complex and simple sentence pairs. [sentence-level]
\end{itemize}

The overall calculation of the simplification processes in the “Saaq al-Bambuu” corpus illustrated in figure 1. The $REWRITE$ operation has the highest proportion of the simplification processes [keeping the word as it is in both versions] in which 21899 words were copied in the simplified version. Whereas, 12561 words have been deleted to simplify the sentence that annotated with $DELETE$ label. In the third position comes $REPLACE$ operation in which 9082 words where subsisted with their simple counterparts. At last, only 362 words were added to simplify the sentences that annotated with $ADD$ label. 

\begin{figure}[!h]
\begin{center}

\includegraphics[scale=0.40]{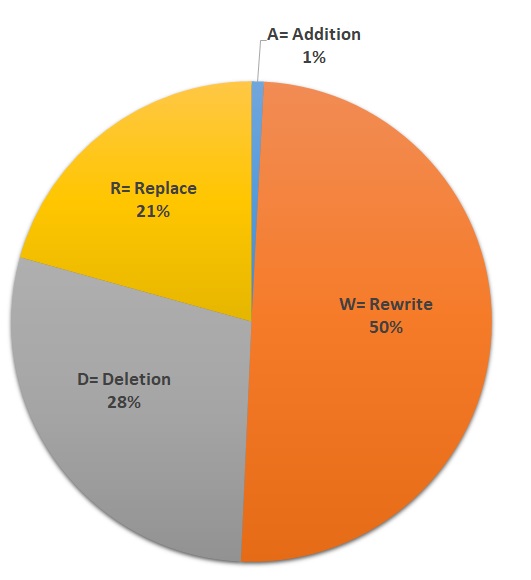} 

\caption{Represents the percentage of each simplification operation on Saaq al-bambuu corpus}
\label{fig.1}
\end{center}
\end{figure}

Regarding Part-Of-Speech features (POS-features) extraction we used MADAMIRA a robust Arabic morphological analyser and part of speech tagger \cite{pasha2014madamira}.

\section{Method One - Classification approach}

The reference for this approach is the pipeline of the LS task, that focuses on LS by replacing complex vocabularies or phrasal-chunks with suitable substances \cite{paetzold2017survey}. To reach this goal, we decided to implement three classification models:
\begin{enumerate}
    \item classification model which is based on \emph{word embedding}, thus we applied \emph{\textbf{fastText}} 
    \footnote{\url{https://fasttext.cc/docs/en/}}
    word embedding tool that represents words as vectors embedding.Those vectors embedding was trained on Common Crawl and Wikipedia. We used the Arabic ar.300.bin file in which each word in WE is represented by the 1D vector mapped of 300 attributes\cite{grave2018learning}; 
    \item{classification model which is based on \emph{transformers}. Using \emph{\textbf{Arabic-BERT}}\footnote{\url{https://huggingface.co/asafaya/}} a  pre-trained transformer model on both filtered Arabic Common Crawl and a recent dump of Arabic Wikipedia  contain approximately 8.2 Billion words \cite{safaya2020kuisail} ;}
    \item{classification model combining both \emph{fastText} and \emph{Arabic-BERT} results with post-editing rules;}
    \end{enumerate}

Considering the definition of the four main steps applied in the  pipeline for LS as follows:

{\emph{Complex word identification [CWI]}} is the main first step performed at the top of the pipeline that employed to distinguish complex words from simple words in the sentence.
{\emph{Substitution Generation [SG]}} involves generating all possible substitutions but without including ambiguous substances that would confuse the system in the Substitution Selection step.
{\emph{Substitution Ranking [SR]}} is to order the new generated substitution list to ease the selection step by giving high probability of the most appropriate highly ranked word.
{\emph{Substitution Selection [SS]}} is responsible for selecting from the ordered SG’s generated list the most appropriate substitute according to the context while preserving the same meaning and grammatical structure.Taking into account the fact that, a word may have multiple meanings, and different meanings will have different relevant substitutions, then the SS task may generate a miss-substitution, which may lead to meaning corruption.
The following part of this paper moves on to describe in greater detail the implementation of each step concentrating on employed methods and tools.
\subsection{Complex word identification}
CWI step could be viewed as a layered analysis opt for a better understanding of word complexity. Hence, we applied a lexicon-based approach. Taking into account one sentence per time, the first level relates to identifying the number of syllables per each word in the target sentence keeping record of its POS-tag along with other features produced by MADAMIRA to be used in further steps. The second layer of analysis moved to assign each word a CEFR complexity level adopting a Lexical based approach using CEFR vocabulary List
as a reference to allocate each word in the target sentence to a readability level.
At CWI, with identifying the complex words, these words become the targets to simplify. Unfortunately, it is impractical to simplify all complex words in a sentence at once so that, first by ordering words according to their CEFR level and taking into account each of these words as the target per time to deploy the simplification process. For example, if a sentence has three complex words assigned with B2, C2, C1, firstly we order them to be C2, C1, and B2 and then start the simplification process with targeting C2 tagged word, followed by C1 and so on. In this example, this operation results in generating three sentences each with different masked word slot.
\subsection{Substitution Generation and Ranking}
These two steps were considered in one process using different methodologies to generate the substitution list and ranking them considering semantic similarity measures. For this purpose we obtained different sentence embedding to produce ten top ranked substitution list of the masked token.


\subsubsection{Arabic-BERT prediction}
Firstly, for each \emph{complex word} we use Arabic-BERT model with applying BERT’s task $Masked Language Modeling$ (MLM). This task predicts a substitution list of a masked [not shown, complex] token in a sequence given its left and right context. At this process, the MLM requires a concatenation between the original sequence and the same sentence sequence where the target word is replaced by [MASK] token as a sentence pair, and feed the sentence pair into the BERT to obtain the probability distribution of the possible replacements corresponding to the MASK word. To use any pre-trained BERT model, we need to convert the input data (sentence’s tokens) into an appropriate format so that each sentence can be sent to the pre-trained model to obtain the corresponding embedding using modules and functions available in Hugging Face’s transformers package. For this task, first, convert tokens’ vectors to $PyTorch$ tensors. Secondly in the next sentence prediction, the beginning and end of each sentence need to be marked before feeding them to the BERT model. For this purpose, a general token [CLS] was added as a first token to represent the hidden state of the whole sentence along with adding another generated token [SEP] identifying the end of a sentence. For example, any input could be represented by: [CLS] original sentence [SEP] sentence with a masked token [SEP], in which [CLS] is the beginning of the sentence, the first [SEP] a mark for the end of the first sentence and the beginning of the following one and, a last [SEP] identifying the end of the whole input. By using this approach, taking into account not only considering the complex word, but also the surrounded context of the complex word. \textbf{For example}, given this sentence from Arabic Wikipedia:
\begin{RLtext}
 tata.talab min hay'aT alma.hkamaT wujwub ta.hdiyd al.huquwq
\end{RLtext}
{\it{tata\d{t}alabu min hay'atu alma\d{h}kamatu wujūba ta\d{h}dīda al\d{h}uqūq}}

[But many situations require the court to determine the rights]

\begin{figure}[!h]
\begin{center}
\includegraphics[scale=0.35]{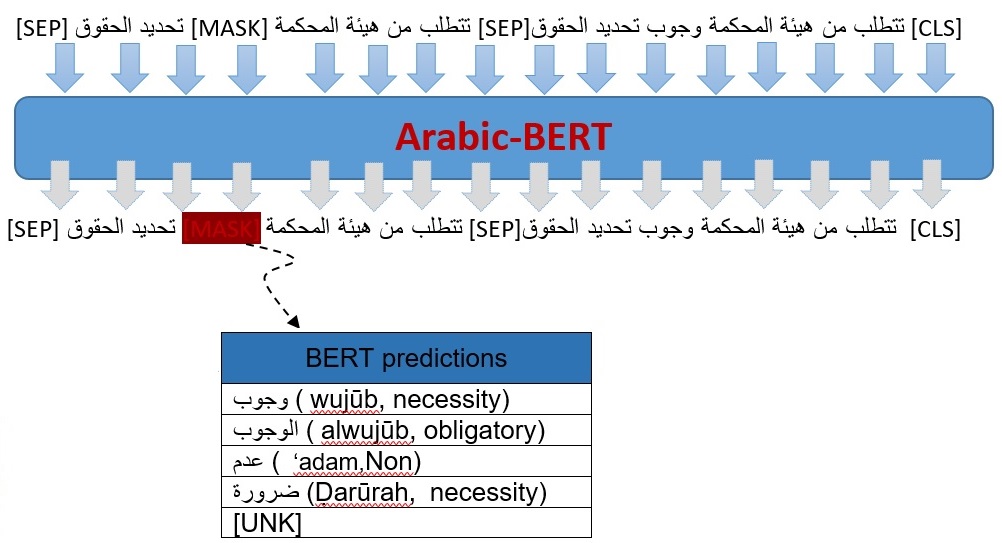} \caption{The complex word in this sentence is {\it“Wujūba”}  (’ \RL{wujwub}’, ‘necessity’), to get the simplest replaces candidates we will feed it into Arabic-BERT}
\label{fig.2}
\end{center}
\end{figure}

The sentence pair construction before feeding into Arabic-BERT is shown in figure \ref{fig.2}. Also this figure illustrate part of the prediction list of the [MASK] word {\it“Wujūba”}  (’ \RL{wujwub}’, ‘necessity’) applying MLM task [BertForMaskedLM] from the hugginface library.

One of the most noticeable aspect in BERT is sentence tokenization that is an initial step before converting tokens into their corresponding unique IDs [embedding vector]. There is an important point to highlight about BERT-tokenizer algorithm, the common out-of-vocabulary (OOV) problem, since the model is pre-trained on a specific corpus, the words are limited to ones that appeared in this training corpus. As a solution, in testing and prediction processes, BERT models is designed to replace the unseen tokens with a special token [UNK], which stands for unknown token. 

However, converting all unseen tokens into [UNK] will take away a lot of information from the input data. Hence, the BERT tokenizer adopts the $WordPiece$ algorithm that not only splits the sentences into words but also breaks out words into several subwords. This splitting technique is represented by the model by adding ‘\#\#’ as a start for each consecutive word part. The BERT tokenization function, on the other hand, will first split the word tata\d{t}alabu (’ \RL{tata.talab}’, ‘require’) into two subwords, namely [\RL{tata}'] and ['\RL{.talab}\#\#’], where the first token is a more commonly-seen word (prefix) in a corpus, and the second token is prefixed by two hashes \#\# to indicate that it is a suffix following some other subpart. If there is no way how to split the token into subwords, the whole word becomes [UNK]. After this tokenization step, all tokens can be converted into their corresponding IDs.

\subsubsection{fastText prediction}
\begin{figure}[!h]
\begin{center}

\includegraphics[scale=0.38]{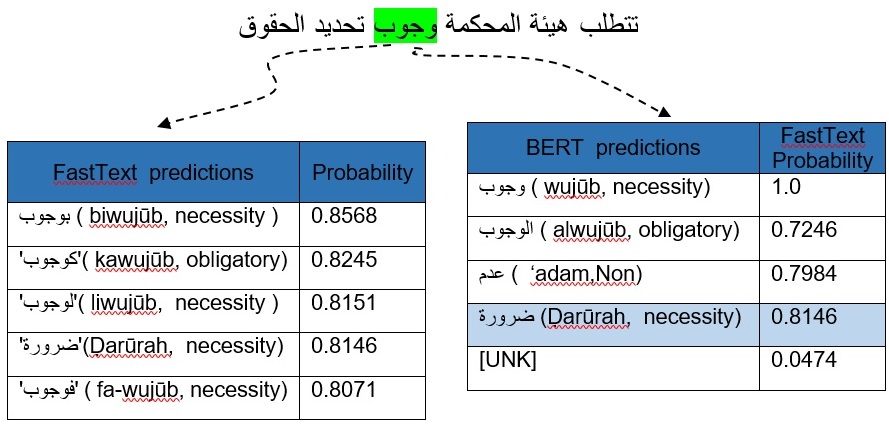} 

\caption{Arabic-BERT and fastText predication lists along with the probability obtained from fastText for the word “Wujūba”  (’\RL{wujwub}’, ‘necessity’)}
\label{fig.3}
\end{center}
\end{figure}

Using \emph{fastText} model in two folded processes, first ranking the previously produced substitutions obtained by MLM BERT. This is done by calculating the semantic cosine similarity between each word in the produced list to the target complex word. The second process is using fastText word embedding itself to generate a list of possible replacements [SG] and then ranking by the nearest neighbour [SR]. For example, the fastText generated list given the target complex word in the previous example is shown on the left side of the figure \ref{fig.3}. Whereas, the ranking probability of Arabic-BERT's prediction list using fastText   was shown on the right side of figure \ref{fig.3}.

\subsection{Substitution Selection}
At this stage, each complex word in the sentence has different ordered substituted lists based on Arabic-BERT and fastText. Taking into account each prediction list to analyse individually and select the more logical substitute based on the probabilities and some linguistics rules. This allowed the system to generate a set of simplified versions of the target sentence. In addition, keeping a record of the semantic similarity and the readability level of the new produced sentences. The system produces three different simple sentences based on Arabic-BERT substitute selection, fastText, and Combined decision from both generated lists.
The combined decision is a very crucial stage and the system needs to be careful when selecting the best substitute based on different measures. Starting with the Arabic-BERT list, the greater the value the most common or familiar is the word for a person referring to simple words. If the word is tagged with replacement with [UNK] the decision is to ignore the results from Arabic-BERT and rely on fastText results. Then, applying the following four rules to limit incorrect selection:
\begin{enumerate}
  \item \texttt{Rule1:} if [UNK] is a top-ranked substitute then go to fastText results.
  
Check if the first substitute is [UNK] in this case the system completely ignores BERT results and keep the original then rely on FastText results immediately.

  \item \texttt{Rule 2:} if any word’s lemma in the generated list equal the lemma of the original word excludes these words from the list.
  
Check if the lemmas in the predicted list matches the same lemma of the target word. In this case, we exclude these words from the potential replacement for the target word and keep only the words with a different lemma. These replacements should also share the same POS and Number with the target word. 

  \item \texttt{Rule 3:}if the substitute word is more difficult than the target word.
  
Check the word CEFR level of the new substitute word. The new word's CEFR level should be equal to or less than the CEFR level of the target word. Because sometimes the generated list may have a more frequent substitute which is more difficult than the original word but more frequent. 
  \item  \texttt{Rule 4:} check if the new substitute shares the meaning. 
  
The system use this rule as it gives a level of confidence to the system selection. After the system makes the final decision either, keep the target word or select the suggested substitute based on previous rules. At this stage, comparing both target and substitute MADAMIRA English translation feature [appeared in Gloss feature]. If both words share part or all possible translation this gives the system confidence to replace the target with the substance. 
\end{enumerate}

\section{Method Two: Generative Approach}

Here, we employ a Seq2Seq approach adopting  \textbf{\emph{T5 “Text-to-Text Transfer Transformer”}}\footnote{\small{\url{https://simpletransformers.ai/docs/t5-specifics/}}}. T5  is a BERT-like transformer that takes input a text and training it on the model to generate target text of a different variety of NLP text-based tasks such as (summarization, translation, question answering and more) \cite{raffel2019exploring}. The main difference between BERT and T5 is that BERT uses a Masked Language Model (MLM) and an encoder-decoder, although T5 employs a unified Seq2Seq framework \cite{farahani2021parsbert}. T5 model initially targeted English-Language NLP tasks. Recent research extended the model to include more than 101 languages including the Arabic Language. A “multilingual Text-to-Text Transfer Transformer”, Multilingual T5, mT5 \cite{xue2020mt5}, a new variant of T5 and pre-trained on Common Crawl-based dataset. The pre-trained language model was very successful for the Natural Language Understanding (NLU) task. 


Considering the multilingual capabilities of mT5 and the suitability of the Seq2Seq format for language generation. This gives it the flexibility to perform any NLP task without having to modify the model architecture in any way. This experiment employs the ‘MT5-For-Conditional-Generation’ class that is used for language generation. Training a TS model using "Saaq  al-Bambuu" parallel sentences, over the mT5-base model\footnote{google/mt5-base, is available through Huggingface repository, \small{\url{https://huggingface.co/google/mt5-base}}}. The system was developed in $Python 3.8$ environment with using other toolkits such as Natural Language Processing Toolkit ($NLTK$ \footnote{\small{\url{http://www.nltk.org/}}}) and $Scikit-learn$\footnote{\small{\url{https://scikit-learn.org/stable/index.html}}}. Our sentence corpus was randomly split into 80\% for training and 20\% for testing. 

\section{Evaluation}
Likewise, most TS evaluation approaches have been driven from other similar NLP research areas. Various evaluation methods have been applied across researches to measure the three main aspects of the newly generated text. These aspects are, i) fluency, referring to the grammatically well-formedness and structure simplicity; ii) adequacy, meaning preservation; iii) simplicity, more readable. All methods were evaluated on the same test dataset that consisted of 299 randomly chosen sentences excluded from training. We employed both automatic and manual evaluation comparing both systems.
\subsection{Automatic Evaluation}
BERTScore is an evaluation metric that computes cosine similarity scores using BERT-style embedding from a pre-trained transformer model. As such models provide a better representation of the linguistic structure, BERTScore evaluation correlates better with human judgments regarding the measurements of sentence similarity. BERTScore evaluation metric overcome the limitations of the previous Machine translation evaluation metrics such as BLEU\cite{papineni2002bleu} and SARI\cite{xu2016optimizing}, n-gram based evaluation metrics. These methods were not able to capture two main simplification features: 1) changing word order as paraphrasing simplification method, 2) maintaining the deep structure meaning, despite changes in the surface form structure. Moreover, the BERTScore evaluation method gives the option to use different pre-trained transformer models by applying \textit{baseline rescaling} to adjust the output scores.  This allowed determining the performance of different Arabic-language trained BERT models;(i) the default in multilingual BERT (mBERT)\cite{devlin2018bert} that is based on the selected language which is Arabic in this case; (ii) ARBERT\footnote{\small{\url{https://github.com/UBC-NLP/arbert}}}, that trained on a collection of six Arabic datasets comprising 61GB of text (6.2B tokens) \cite{abdul-mageed-etal-2021-arbert}; (iii)AraBERTv0.2-base \footnote{\small{\url{https://huggingface.co/aubmindlab/bert-base-arabert}}} model consist of 77GB of sentences (8.6B tokens) \cite{antoun2020arabert}.   However, AraBERT has been trained on a larger corpus than ARBERT, the latter uses WordPeice tokeniser as illustrated before. Whereas, AraBERT relies on SentencePiece tokeniser that uses spaces as word boundaries. Considering these two parameters reflected in BERTScore metrics.

\paragraph{\emph{Classification approach - Automatic Evaluation}}

\begin{figure}[!h]
\begin{center}

\includegraphics[scale=0.6]{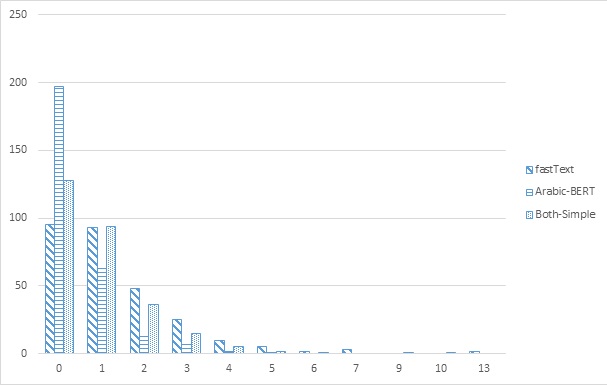} 
\caption{number of changed words using fastText-alone, Arabic-Bert-alone and combined}
\label{fig.5}
\end{center}
\end{figure}

The classification system produced three simple versions of the target sentence using BERT-alone, fastText-alone, and combined version. This automatic evaluation was applied to compare different BERT models resolutions of these sentences as represented in Table 2. Figure \ref{fig.5} represents the number of changes performed by each classification model. These primarily results suggests that using fastText-alone perform unneeded simplification resulting in lower F-1. Whereas, a higher F-1 measure in Arabic-BERT-alone generated sentence suggest that using BERT eliminate necessary changes. While the combination of both tools suggestions enhances the substitution ranking and choice process. That eliminates unnecessary changes and enhance performance. In this case, combined produced sentences achieved P 0.97, R 0.97 and F-1 0.97 using ARBERT.

\begin{table*}[!t]
\begin{center}

\begin{tabular}{cc}

\begin{tabular}{|l|lll|}
\hline
\textbf{Classification} & \multicolumn{1}{l|}{\textbf{P}} & \multicolumn{1}{l|}{\textbf{R}} & \textbf{F1} \\ \hline
                   & \multicolumn{3}{l|}{Default mBert  }                            \\ \hline
Target/fastText    & \multicolumn{1}{l|}{0.962}      & \multicolumn{1}{l|}{0.966}      & 0.964       \\ \hline
Target   /BERT     & \multicolumn{1}{l|}{0.991}      & \multicolumn{1}{l|}{0.990}      & 0.\textbf{990}       \\ \hline
Target / Combined  & \multicolumn{1}{l|}{0.974}      & \multicolumn{1}{l|}{0.975}      & 0.975       \\ \hline
                   & \multicolumn{3}{l|}{ARBERT}                                             \\ \hline
Target/fastText    & \multicolumn{1}{l|}{0.958}      & \multicolumn{1}{l|}{0.960}      & 0.959       \\ \hline
Target   /BERT     & \multicolumn{1}{l|}{0.990}      & \multicolumn{1}{l|}{0.991}      & 0.\textbf{990}       \\ \hline
Target / Combined  & \multicolumn{1}{l|}{0.976}      & \multicolumn{1}{l|}{0.976}      & 0.978       \\ \hline
                   & \multicolumn{3}{l|}{AraBERT}                               \\ \hline
Target/fastText    & \multicolumn{1}{l|}{0.962}      & \multicolumn{1}{l|}{0.963}      & 0.963       \\ \hline
Target   /BERT     & \multicolumn{1}{l|}{0.989}      & \multicolumn{1}{l|}{0.989}      & 0.989       \\ \hline
Target  / Combined & \multicolumn{1}{l|}{0.975}      & \multicolumn{1}{l|}{0.976}      & 0.976       \\ \hline
\end{tabular}
     & 
\begin{tabular}{|l|lll|}
\hline
\textbf{Generation} & \multicolumn{1}{l|}{P}       & \multicolumn{1}{l|}{R}       & F1     \\ \hline
                    & \multicolumn{3}{l|}{Default mBert }                 \\ \hline
Original/Target     & \multicolumn{1}{l|}{0.889}   & \multicolumn{1}{l|}{0.838}   & 0.862  \\ \hline
Generated/Original  & \multicolumn{1}{l|}{0.806}   & \multicolumn{1}{l|}{0.725}   & 0.762  \\ \hline
Generated/   Target & \multicolumn{1}{l|}{0.754}   & \multicolumn{1}{l|}{0.723}   & 0.736  \\ \hline
                    & \multicolumn{3}{l|}{ARBERT}              \\ \hline
Original/Target     & \multicolumn{1}{l|}{0.840}   & \multicolumn{1}{l|}{0.754}   & 0.790  \\ \hline
Generated/Original  & \multicolumn{1}{l|}{0.647}   & \multicolumn{1}{l|}{0.529}   & 0.573  \\ \hline
Generated/   Target & \multicolumn{1}{l|}{0.570}   & \multicolumn{1}{l|}{0.524}   & 0.538  \\ \hline
                    & \multicolumn{3}{l|}{AraBERT} \\ \hline
Original/Target     & \multicolumn{1}{l|}{0.879}   & \multicolumn{1}{l|}{0.823}   & 0.848  \\ \hline
Generated/Original  & \multicolumn{1}{l|}{0.787}   & \multicolumn{1}{l|}{0.693}   & 0.734  \\ \hline
Generated/   Target & \multicolumn{1}{l|}{0.723}   & \multicolumn{1}{l|}{0.686}   & 0.701  \\ \hline
\end{tabular}

\end{tabular}

\caption{Precision, recall and F1 measures using BERTScore with different transformer models}

\label{tabEvaluation}
\end{center}
\end{table*}

\paragraph{\emph{Generative Approach-Automatic Evaluation}}
Testing the 299 sentences for evaluating the generated simplified sequences compared to the original sentences and the target simple sentences. Using three measures as presented in Table 2:
\begin{itemize}
    \item{	Original/Target, considering it as a reference to the mT5 system. }
    \item{	Generated/Original, comparing the newly generated sentence with the original complex sentence.}
    \item{	Generated/Target, comparing the newly generated sentence with the target simple sentence.}
   
 \end{itemize}

\begin{figure}[!ht]
\begin{center}

\includegraphics[scale=0.36]{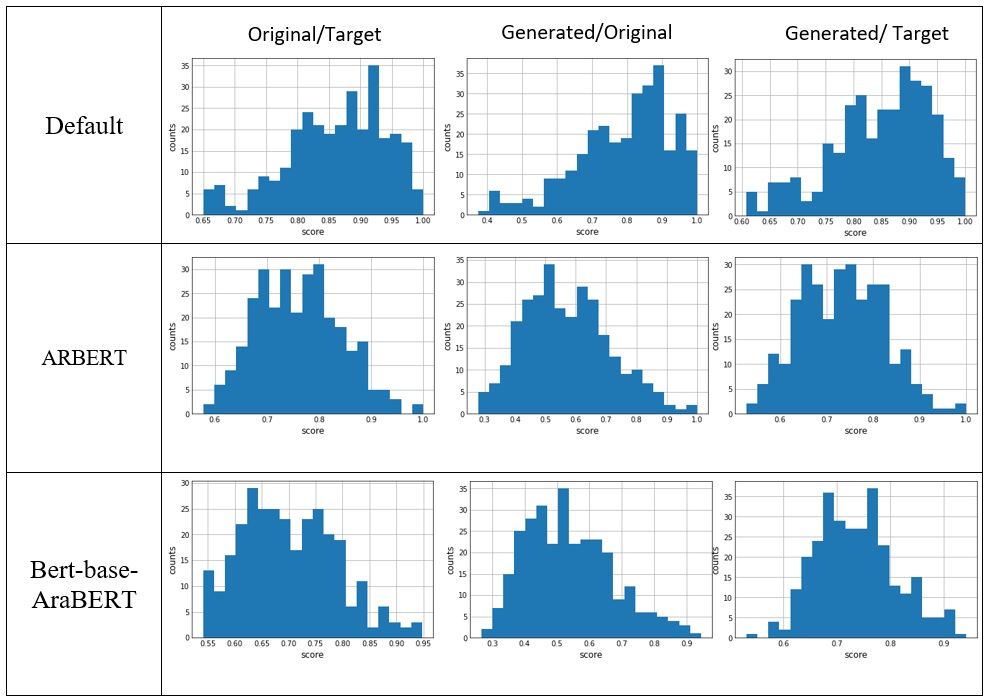} 

\caption{The F1 scores for each sentence pair, the scores are more spread out, which makes it easy to compare different methods}
\label{fig.6}
\end{center}
\end{figure}

To further illustrate these three models’ performance, figure \ref{fig.6}, represents the distribution of F-1 across the testing data instances using different BERT models. The default model F-1 plots skewed towards the right reflecting strong similarity across the three parallel sentences (Original/Target/Generated). Whereas, AraBERT plots Original/Target and Generated/Original skewed to the left indicating less similarity across the data. While, ARBERT’s plots represent a normal distribution representing a more accurate similarity measure in the data. This findings suggests ARBERT that applying a WordPeice sentence tokeniser BERT model performed better in sentence representation.

\subsection{Manual Evaluation}
\begin{figure}[!ht]
\begin{center}
\includegraphics[scale=0.6]{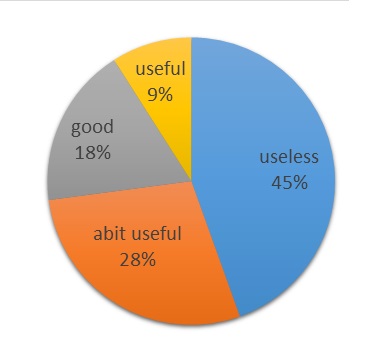} 
\caption{Simplified sentences analysis based on the usefulness of the lexical substitution processes.}
\label{fig.7}
\end{center}
\end{figure}

\textbf{\emph{Classification Approach - Manual Evaluation}} a manual analysis of the produced sentences of combined system has been performed. The results displayed in figure\ref{fig.7} on a scale of good, useful, a bit useful, and useless simplification. 55\% of the new simplified sentences were either good, useful or a bit useful as a majority. While 45\% of the sentences were classified as useless simplification where the complex word was replaced either by a more complex word or its antonym. For example, a useful simplification from the combined system as in this sentence from "Saaq al-Bambuu",
\begin{RLtext}
kuntu '.hadq fiy Al.tabaq wAl.samt yakAdo yabtal` AlmakaAn
\end{RLtext}
{\it{Kuntu 'u\d{h}addiqu fī al\d{t}abaqi wa-al-\d{s}amtu yakādu yabtali` al-makān.}}

[I was staring at the plate and the silence almost swallowed up the place.]

In this sentence, the word ‘\RL{'.hadq}’  ({\it{U\d{h}addiqu}},’ staring’) was replaced by \RL{'ta'mal} ( ‘ata'mmalu, ‘muse’), that is more frequent and simpler and generate:

\begin{RLtext}
kuntu 'ta'malu fiy Al.tabaq wAl.samt yakAdu yabtal` AlmakaAn
\end{RLtext}
Although, it is simpler it doesn't reach the exact target word \RL{'n.zur} ( ‘An\d{z}uru, ‘look’)

\begin{figure}[!ht]
\begin{center}

\includegraphics[scale=0.72]{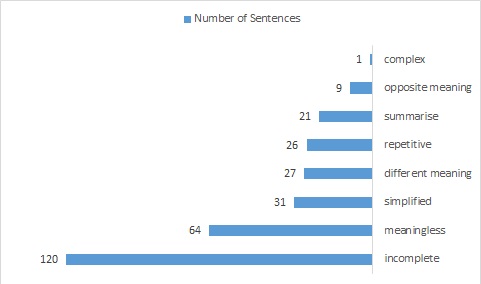} 

\caption{Manual error analysis distribution across testing data}
\label{fig.8}
\end{center}
\end{figure}

\textbf{\emph{Generative Approach-Manual Evaluation}}
despite the initial automatic evaluation provided promising results, the manual evaluation of the generated text provides deeper insight into mT5’s output for the Arabic simplification task. According to the manual error analysis as shown in figure \ref{fig.8} only 31 sentences were correctly simplified from 299 testing instances. In addition, about 120 generated sentences were incomplete and the system produced 64 meaningless or ill-formed sentences. A significant shortcoming that the produced sentences tends to have the same repeated phrase. Moreover, one of the generated sentences were more complex than the original sentence. Also, the unexpected errors were producing simple sentences, yet with different or opposite meanings. As such example giving an opposite meaning in the generated sentence:
\begin{RLtext}
`lAqaty bAlkaniysT fiy bilAd 'umy qawyT jdaN
\end{RLtext}
{\it{'ilāqti bil-Kanīsatu fī bilād 'ummī qawīyya\d{h} jiddan}}
That means, [My relationship with the Church in my mother's country is \textbf{very strong}.] This simple version contradict the meaning of the original sentence.
\begin{RLtext}
lays hunAk mA yumayz `lAqaty baAlkaniysT fiy blAd 'umy fzyArAty lahA qalylaT jdaN
\end{RLtext}
{\it{Laysa hunāk mā yumayyizu 'ilāqati bil-Kanīsatu fī bilād 'ummī, faziyārāti lahā qalīlah jiddan}}
[There is nothing that distinguishes my relationship with the Church in my mother's country. \textbf{My visits to it were very few}.]

In this sentence, instead of mentioning that his relationship was not strong, the system generated the opposite meaning by expressing a very strong relationship. 

Otherwise, mT5 in some cases can produce a perfectly valid paraphrase, which is better than the target simple sentence.
\begin{RLtext} \textbf{.talab mnA Aljlows} fiy .saAlownah  almaly' biAlkotub \end{RLtext}
{\it{\d{t}alab minnā al-julūs fī \d{s}ālwnahu almaly' bi-al-kutub}}

[He asked us to sit in his salon full of books.]
\begin{RLtext} fiy .saAlownah Al.sa.gyr almaly' biAlkotub \textbf{.talab mnA Aljlows} 'mAm maktab .sa.gyr\end{RLtext}
{\it{Fī \d{s}ālunahu al-\d{s}aghīr almali' bil-kutubi, \d{t}alaba minnā al-julūsi 'amāma maktabi \d{s}aghīri}}

[In his small salon full of books, he asked us to sit in front of a small desk.]
In this case, the generated sentence was syntactically simpler than the target while focusing on the main information.

\section{Related Works}

\newcite{blum1978universals} completed one of the first studies that introduce Lexical simplification for Teaching English as a Second Language (TESOL). Some of the following TS systems applied a rule-based approach \cite{petersen2007text,evans2014evaluation}. Most later carried out studies  based on a monolingual parallel-aligned corpus of original and simplified texts by applying different machine-learning algorithms such as \newcite{aluisio2008towards} and \newcite{caseli2009building} for Portuguese language, \newcite{collados2013syntactic} for Spanish language and \newcite{glavavs2015simplifying} for English. Other researchers considered the TS problem as a monolingual translation problem that is best solved through applying the Statistical Machine Translation (SMT) framework  \cite{specia2010translating,zhu2010monolingual,woodsend2011wikisimple,wubben2012sentence}. Latest English TS studies start applying word embedding\cite{paetzold2016semeval,paetzold2017lexical} and BERT transformers for lexical simplification as presented in {\newcite{qiang2020lexical}} proving its effectiveness in solving LS task. 

Unlike English and Other Latin languages, only a few researchers have been tackling the problems of Arabic ATS. 
\newcite{al2011baseet} a prototype unreleased system at King Saud University, they proposed Arabic Automatic Text simplification system (AATS) called Al-basset. The system architecture for AATS structured in the light of the state of the art of systems for other languages. Such as  SYSTAR, a syntactic simplification system for the English aphasic or inarticulate population\cite{carroll1998practical}. Another system, SIMPLIFICA, is a simplification tool for Brazilian Portuguese (BP) targeting those with low literacy levels \cite{scarton2010simplifica}. The design of "Al-Basset" was constructed of four main stages: i) measuring complexity, in this stage they would adopt a statistical language model based on a machine learning technique called ARABILITY \cite{al2010automatic}; ii) vocabulary (lexical) simplification by following the LS-pipeline and produce the synonyms either by  building a new dictionary or using Arabic-WordNet\cite{rodriguez2008arabic} while select the most common and possible synonym, by using the Google API; iii) syntactic simplification, they suggested identifying the complex structures by applying a look-up approach to a manually predefined list of Arabic complex structures; iv) diacratization using MADA\cite{habash2009mada+} diacritizer task. The main limitation of implementing this system at this point is the unavailability of Arabic basic resources and tools. Such as dictionaries, corpora and parallel complex-simple structures which are the main components of any ATS system.

\newcite{al2017simplification} provided the second attempt to build an AATS system at New York University in Abu-Dhabi. Their simplification system was designed to be semi-automatic to simplify Arabic modern fiction; it involved a linguist using a web-based application to apply ACTFL (American Council on the Teaching of Foreign Languages) language proficiency guidelines for simplification of five Arabic novels. They aimed to provide essential Arabic resources for building ATS and formulating manual simplification rules for Arabic fiction novels using TS stat-of-the-art.  The first resource they expected to produce is a  corpus consisting of 1M tokens of the 12-grade curriculum, 5M tokens of the adult novels (original and simplified counterparts), and 500K tokens of children’s stories. Also, they provided a proposal to the SAMER (Simplification of Arabic Masterpieces for extensive reading) project based on the corpus analysis. Their guidelines invoke both the MADAMIRA \cite{pasha2014madamira} and CAMAL dependency parser \cite{shahrour2016camelparser} for data analysis and classification of their corpus. They were aiming to build a readability measurement identifier to formulate a 4-levelled graded reader scale (GRS) by applying various machine-learning classifiers. 

It should be noted that, neither of them followed up with any further application of their success or failure.

\section{Conclusion}
In this paper, we have presented the first Modern Standard Arabic sentence simplification system by applying both classification and generative approaches. On the one hand, the classification approach focuses on lexical simplification. We looked at the different classification methods and we found that a combined method generates well-formed simple sentences. In addition, using word embeddings and transformers prove to produce a reasonable set of substitutions for the complex word more accurately than traditional methods such as WordNet. Our interpretation of the limitation of the classification system arises from the fact that some of the generated sentence structures are not well-formed and that the system can misidentify what makes some complex words in the CWI step. Even though this limitation reveals the limitations of the Arabic CEFR vocabulary list in identifying the complex word, the list is shown to be more useful in the substitution replacement step.
On the other hand, while the generative Seq2Seq approach provides a less accurate simplified version in most cases, in some cases it outperforms the classification approaches by generating a simplified sentence, which can be even better than the target human simple sentence. Nevertheless, one of the limitations of the generative approach can be the repetition of a part of the same phrase patterns. Future research is needed to address this issue.

Overall, we show the advantages and limitations in the two approaches, both of which could benefit from building a larger parallel simple/complex Arabic corpus. Moreover, adding a post-handler language generation module could resolve some of the limitations even if only acting as a less accurate alternative fast solution, for example, by avoiding and removing repeated phrase patterns produced from the generative system. Even though Arabic Text Simplification is a very challenging task, our research demonstrates huge potential towards achieving a better developed system even starting from a small corpus.

\section{Acknowledgement}

This research is a part of PhD project funded by Newton-Mosharafa Fund. All experiments presented in this paper were performed using Advanced Research Computing (ARC) facilities provided by Leeds University.

\section{Bibliographical References}\label{reference}

\bibliographystyle{lrec2022-bib}
\bibliography{lrec2022-example}

\bibliographystylelanguageresource{lrec2022-bib}
\bibliographylanguageresource{languageresource}

\end{document}